\documentclass[table]{fairmeta}
\usepackage{wrapfig}
\usepackage{textcomp}
\usepackage{stfloats}
\usepackage{url}
\usepackage{verbatim}
\usepackage{titlesec}
\usepackage{tocloft}
\usepackage{adjustbox}
\usepackage{pifont}
\usepackage{tikz}
\usepackage{comment}
\usepackage{amsmath,amssymb} % define this before the line numbering.
\usepackage{graphicx}
\usepackage{float}
\usepackage{algorithm}
\usepackage{listings}
\usepackage{hyperref}

\usepackage{array}       % 必须在 colortbl/xcolor 之前
\usepackage{tabularx}
\usepackage{multirow}
\usepackage{booktabs} 

\usepackage{subcaption}  % 用于 subfigure 和 subtable

\usepackage{bbding}      % \Checkmark,\XSolid,...
\usepackage{fontawesome}

\RequirePackage{xspace}
\makeatletter
\DeclareRobustCommand\onedot{\futurelet\@let@token\@onedot}
\def\@onedot{\ifx\@let@token.\else.\null\fi\xspace}

\makeatother

\definecolor{adptorange}{RGB}{248, 205, 172}
\definecolor{cmpblue}{RGB}{189, 215, 238}
\definecolor{our_red}{RGB}{232,157,160}
\definecolor{our_blue}{RGB}{136,206,230}
\definecolor{our_orange}{RGB}{246,200,168}
\definecolor{our_green}{RGB}{178,211,164}

\definecolor{attn_code0}{RGB}{247,215,200}
\definecolor{attn_code1}{RGB}{238,169,139}
\definecolor{mlp_code0}{RGB}{204,201,221}
\definecolor{mlp_code1}{RGB}{102,95,153}

\definecolor{token_blue}{RGB}{84, 120, 140}
\definecolor{codeblue}{rgb}{0.25, 0.5, 0.5}
\definecolor{codekw}{rgb}{0.35, 0.35, 0.75}
\definecolor{green}{HTML}{009000}
\definecolor{red}{HTML}{ea4335}

\newlength\savewidth

\newcolumntype{x}[1]{>{\centering\arraybackslash}p{#1pt}}
\newcolumntype{y}[1]{>{\raggedright\arraybackslash}p{#1pt}}
\newcolumntype{z}[1]{>{\raggedleft\arraybackslash}p{#1pt}}

\renewcommand{\paragraph}[1]{\vspace{1.25mm}\noindent\textbf{#1}}
\lstdefinestyle{Pytorch}{
    language = Python,
    backgroundcolor = \color{white},
    basicstyle = \fontsize{9pt}{8pt}\selectfont\ttfamily\bfseries,
    columns = fullflexible,
    aboveskip=1pt,
    belowskip=1pt,
    breaklines = true,
    captionpos = b,
    commentstyle = \color{codeblue},
    keywordstyle = \color{codekw},
}

\title{A Contrastive Pre-trained Foundation Model for Deciphering Imaging Noisomics across Modalities}

\author[* 1,2]{Yuanjie Gu}
\author[* 1,3]{Yiqun Wang}
\author[* 4,5]{Chaohui Yu}
\author[1]{Ang Xuan}
\author[4]{Fan Wang}
\author[\dagger 2]{Zhi Lu}
\author[\dagger 1]{Biqin Dong}

\affiliation[1]{College of Biomedical Engineering, Yiwu Research Institute, Fudan University, Shanghai, China\\}
\affiliation[2]{Department of Psychological and Cognitive Sciences, Tsinghua University, Beijing, China\\}
\affiliation[3]{Shanghai Innovation Institute, Shanghai, China\\}
\affiliation[4]{DAMO Academy, Alibaba Group, Beijing, China\\}
\affiliation[5]{Hupan Laboratory, Hangzhou, Zhejiang, China}

\contribution[*]{Equal contribution}
\contribution[\dagger]{Corresponding author}

\abstract{
Characterizing imaging noise is notoriously data-intensive and device-dependent, as modern sensors entangle physical signals with complex algorithmic artifacts. Current paradigms struggle to disentangle these factors without massive supervised datasets, often reducing noise to mere interference rather than an information resource. Here, we introduce “Noisomics”, a framework shifting the focus from suppression to systematic noise decoding via the Contrastive Pre-trained (CoP) Foundation Model. By leveraging the manifold hypothesis and synthetic noise genome, CoP employs contrastive learning to disentangle semantic signals from stochastic perturbations. Crucially, CoP breaks traditional deep learning scaling laws, achieving superior performance with only 100 training samples, outperforming supervised baselines trained on 100,000 samples, thereby reducing data and computational dependency by three orders of magnitude. Extensive benchmarking across 12 diverse out-of-domain datasets confirms its robust zero-shot generalization, demonstrating a 63.8\% reduction in estimation error and an 85.1\% improvement in the coefficient of determination compared to the conventional training strategy. We demonstrate CoP’s utility across scales: from deciphering non-linear hardware-noise interplay in consumer photography to optimizing photon-efficient protocols for deep-tissue microscopy. By decoding noise as a multi-parametric footprint, our work redefines stochastic degradation as a vital information resource, empowering precise imaging diagnostics without prior device calibration.
}

\date{\today}

\begin{document}
\thispagestyle{firstheader}
\maketitle
\pagestyle{empty}

\section{Introduction} \label{sec:introduction}
\noindent 
Modern imaging systems represent a convergence of physics and computation, yet the stochastic degradations they produce remain poorly understood~\citep{liu2025comprehensive,pei2021effects,zhang2024longterm,chi2025lowdose,vienneau2022spatiotemporal}. While deep learning has revolutionized image reconstruction~\citep{belthangady2019applications,lu2023virtualscanning,lu2025physicsdriven}, the characterization of noise itself has lagged. Noise in contemporary pipelines is no longer a simple Poisson-Gaussian mixture~\citep{zou2025calibrationfree}; it has evolved into a complex, multimodal manifestation of sensor physics, photonic interactions, and black-box algorithmic artifacts. Current approaches lack a universal standard to profile these heterogeneous signatures across diverse modalities without extensive, device-specific recalibration. This gap not only hinders the standardized evaluation of system health, but also precludes the precise optimization of acquisition protocols.

The prevailing paradigm in computational imaging reductively treats noise as a monolithic adversary, an interference to be suppressed in the pursuit of an ideal signal~\citep{mishro2022survey,zhang2024hyperspectral}. This perspective is fundamentally incomplete and perpetuates a significant scientific oversight: It ignores the rich informational content embedded within noise, which serves as a high-fidelity multimodal record of the entire imaging process. Noise carries subtle traces of sensor quantum efficiency, illumination fluctuations, sample-scattering properties, and artifacts introduced by computational processing. Each of these physical and computational interactions leaves a distinct and recoverable imprint on the noise structure, forming a unique multi-parametric fingerprint that is systematically discarded by current methodologies~\citep{chen2008noise}. The consequences of this neglect are substantial, obstructing further progress~\citep{wang2023scientific}.

Contemporary imaging systems integrate components from optical, electronic, and computational domains, leading to the nonlinear accumulation and entanglement of disparate noise signatures~\citep{jiang2026msfa}. Conventional noise models, such as simple Poisson-Gaussian mixtures~\citep{liu2014practical}, fail catastrophically to capture the multi-scale and non-stationary noise behaviors inherent in modern imaging platforms. Although recent physics-based simulations and supervised learning approaches offer limited insights, they remain intrinsically shackled to specific hardware configurations and meticulously curated training data~\citep{zou2025calibrationfree,liu2014practical,zabin1991efficient,escotte1993evaluation}. Their utility diminishes with the introduction of new sensors or imaging modalities, initiating a perpetual cycle of recalibration and retraining. This leaves fundamental questions unresolved, such as how to quantitatively compare noise profiles across diverse imaging devices for standardized diagnostic applications, or how to establish sample-specific excitation parameters that under low-light condition while minimizing photodamage in sensitive biological specimens. These challenges highlight the urgent need for a universal framework for noise analysis, which we term Noisomics. By analogy to genomics, which systematically decodes genetic information to reveal biological function and dysfunction, Noisomics proposes that the systematic decoding of noise components can diagnose imaging system health, quantify subtle biological states, and unlock novel contrast mechanisms previously obscured.

To address this, we propose a universal framework for Noisomics construction and analysis. Central to this approach is the Contrastive Pre-trained (CoP) Foundation Model, which operationalizes the manifold hypothesis~\citep{loaiza-ganem2024deep} to separate high-dimensional signal structures from stochastic perturbations. Unlike supervised methods that rely on scarce paired data, CoP employs a contrastive learning strategy that enforces statistical independence between semantic content and noise features. By projecting inputs into a disentangled latent space, CoP enables the granular quantification of specific noise components, providing interpretable traceability of imaging results.

The CoP framework establishes a robust benchmark for Noisomics, demonstrating superior computational efficiency and unprecedented generalization across twelve out-of-domain (OOD) datasets. By learning a content-independent feature space, CoP transforms noise analysis from passive quality assessment to active, interpretable diagnosis. In consumer photography, we integrate CoP with Explainable AI~\citep{dwivedi2023explainable} to disentangle the non-linear causal links between hardware parameters and noise manifestations. Extending to the scientific domain, we demonstrate CoP’s critical utility in validating photon-efficient protocols for deep-tissue three-photon microscopy. By quantifying the differential reduction of depth-dependent noise components, CoP provides a rigorous statistical verification of depth-gained excitation strategies, offering a measure of efficacy unattainable with traditional aggregate metrics.

\section{Characterizing noise features via CoP learning}
\noindent 
The efficacy of the CoP framework rests on the manifold hypothesis, which posits that clean natural images reside on a low-dimensional semantic manifold, while noise acts as a high-dimensional stochastic perturbation displacing data from this structure (Fig.~\ref{fig1}a). To explicitly characterize these perturbations, CoP employs a contrastive pre-training strategy designed to maximize the mutual information between representations of the same underlying signal while enforcing orthogonality with noise features. Through positive constraints (positive pairs denote two different clean images corrupted by the same parametric noise) and negative constraints (negative pairs denote the same clean image corrupted by different noise), the encoder learns a structured latent space where noise is not treated as mere error, but as a distinct, quantifiable feature vector decoupled from semantic content.

Based on the above assumption, we developed the CoP framework, a two-stage deep learning pipeline designed for the granular, multi-parametric quantification of noise (Fig.~\ref{fig1}b). The framework is anchored by an Expandable Noisomic Engine (ENE), a generative library defining the statistical properties of diverse noise sources. Training begins with an expandable noise genome that systematically corrupts clean images with specific, parametric noise sets drawn from this genome. As visualized in Supplementary Figs.~\ref{figS1}--\ref{figS2}, this engine is capable of simulating both canonical single-source primitives and complex, multi-parametric mixtures across a wide range of intensities, ensuring a comprehensive coverage of the imaging degradation manifold. In this first stage, a contrastive learning for signal agnosticism module forces a deep encoder to generate content-agnostic feature embeddings. By applying a Positive Constraint (P) to pull together features from the same underlying noise genes and a Negative Constraint (N) to push apart features from different noise genes, the model learns a representation space that is exclusively sensitive to the statistical signatures of noise, irrespective of image content. Following this pre-training, in the second stage, the robust encoder is frozen and connected to a lightweight, parametric quantification head, which is then fine-tuned in a supervised manner to precisely estimate the Ground Truth (GT) values of each noise component and the underlying clean signal.

This pre-training strategy delivers substantial performance and efficiency gains. Compared to models trained from scratch, the CoP framework demonstrates significantly accelerated convergence and consistently lower training loss (Fig.~\ref{fig1}c). This advantage in data efficiency holds across all tested training scales, from 100 to over 1 million samples, enabling CoP to achieve a superior validation loss with a fraction of the computational resources (Fig.~\ref{fig1}d). Even the CoP trained with only 100 samples outperformed the traditional method that uses 100,000 samples. Conceptually, the pre-training guides the optimization trajectory towards a more effective minimum in the loss landscape, outperforming solutions achieved by either joint-learning or from-scratch training (Fig.~\ref{fig1}e, f).

\section{Benchmarking and interpretability across diverse imaging modalities}
\noindent 
Benefiting from the noise feature decoupling capability of the latent space, the cornerstone of the CoP framework is its ability for zero-shot generalization across highly diverse, out-of-domain image datasets, a critical failure point for conventional deep learning models. We validated CoP on twelve distinct imaging modalities (Supplementary Fig.~\ref{figS3}), including Chest X-Ray~\citep{irvin2019chexpert}, Histopathology~\citep{wei2021petri}, Retina~\citep{yang2023medmnist}, Alzheimer MRI~\citep{Falah2023Alzheimer}, Microscopy~\citep{kvriza2026Microscopy}, Universe~\citep{liu2020faster}, Satellite~\citep{helber2019eurosat}, Describable Textures~\citep{cimpoi2014describing}, Sketch Scene~\citep{chowdhury2022fscoco}, Manga~\citep{matsui2017sketchbased}, Underwater~\citep{li2020underwater} and Natural Scene~\citep{yang2010image}. Across every dataset, CoP consistently and significantly outperformed models trained from scratch, substantially reducing Root Mean Square Error (RMSE) from 0.058 ± 0.028 to 0.021 ± 0.014 (improving about 63.8\%), and increasing mean Coefficient of Determination ($R^2$) from 0.456 ± 0.458 to 0.844 ± 0.227 in noise parameter estimation (improving about 85.1\%), thus establishing its domain-agnostic generalization capability (Fig.~\ref{fig2}a). Detailed performance breakdown reveals that CoP is particularly transformative for complex noise types: for Poisson, Quantization and Anisotropic noise, CoP improved the classification accuracy by 15.3\%, 55.1\% and 46.7\% compared to the baseline (Supplementary Fig.~\ref{figS4}). Moreover, threshold-accuracy analysis demonstrates that CoP achieves this with superior confidence and slower degradation, maintaining an accuracy of over 22.35\% even at high decision thresholds of 0.30, where baseline performance collapses (Supplementary Fig.~\ref{figS5}).

This remarkable generalization is a theoretical consequence of the feature disentanglement enforced by CoP. A t-SNE visualization~\citep{maaten2008visualizing} of the learned feature space reveals that a model trained from scratch produces embeddings clustered by image content, rendering it domain-specific (Fig.~\ref{fig2}b, left). In stark contrast, the CoP encoder learns a semantically organized latent space where features cluster according to their underlying noise types, regardless of the dataset from which they originated (Fig.~\ref{fig2}b, right). This robust feature clustering is quantified using the Maximum Mean Discrepancy (MMD)~\citep{gretton2012kernel} metric, which shows that the CoP achieved a 21-fold reduction in inter-dataset MMD (0.267 ± 0.198) compared to the baseline (0.013 ± 0.009), confirming it has learned a truly device-agnostic representation essential for Noisomics (Fig.~\ref{fig2}c). Final validation via kernel density estimation~\citep{parzen1962estimation} confirms that CoP’s multi-parametric predictions align tightly with Ground Truth values across all five noise components and the clean signal. Specifically, for the most challenging Anisotropic components, CoP yielded a substantially higher $R^2$ of 0.729, whereas the baseline failed to exceed 0.414 (Fig.~\ref{fig2}d). This high-fidelity characterization validates CoP as a rigorous and reliable tool for quantitative Noisomics.

\section{Disentangling parameter-quality interplay in consumer photography}
\noindent 
The robust, domain-agnostic nature of the CoP framework enables a transition from passive quality assessment to the active, interpretable diagnosis of noise origin in real-world applications. We applied CoP to a large-scale cohort of smartphone images acquired under diverse conditions to systematically understand the complex~\citep{abdelhamed2018highquality}, non-linear dependencies between image acquisition parameters and specific noise components.

First, we validated the model’s fidelity to decouple the underlying “clean signal” by correlating its predictions with established, objective quality metrics. The CoP-blind-estimated clean signal showed a significant positive correlation with Peak Signal-to-Noise Ratio (PSNR) across images captured at both low (ISO 100, mean Pearson’s r = 0.741) and high (ISO 800, mean Pearson’s r = 0.646) sensitivity settings (Fig.~\ref{fig3}a-d), while also maintaining positive correlation with SSIM. This robust correlation confirms that the CoP model accurately disentangles the underlying signal from the accumulated noise, regardless of the camera’s operating point or the level of degradation.

Next, to explore the systematic relationships within the noisomic landscape, we analyzed the interplay between key acquisition parameters (ISO, shutter speed, brightness, smartphone type, and ambient temperature) and the levels of the five predicted noise components. A correlation heatmap provided a preliminary view, revealing a strong positive correlation between ISO and most noise types, while brightness generally exhibited a strong negative correlation (Fig.~\ref{fig3}e). However, these simple linear correlations fail to capture the full, complex, and often non-linear causalities inherent in modern ISPs. To achieve a mechanistic understanding, we leveraged Explainable AI (XAI) techniques~\citep{lundberg2017unified,borys2023explainable,barredoarrieta2020explainable} specifically via the SHapley Additive exPlanations (SHAP) method~\citep{lundberg2017unified}, to quantify the contribution of each parameter to the prediction of individual noise components (Fig.~\ref{fig3}f-j). The SHAP summary plots provide detailed feature attributions: for instance, high ISO values (represented by red points on the right of the plot) consistently contribute positively to the prediction of Gaussian (Fig.~\ref{fig3}f), Poisson (Fig.~\ref{fig3}h), and Quantization (Fig.~\ref{fig3}i) noise, aligned with physical principles where higher sensor gain amplifies electronic and shot noise. Conversely, high brightness values (red points on the left) consistently suppress these noise types. This granular analysis demonstrates the diagnostic power of Noisomics, allowing us to link specific physical causes (e.g., high-gain electronics) to their resulting noise manifestations.

Finally, we synthesized these insights into a Sankey diagram~\citep{riehmann2005interactive,ming2019rulematrix} to visualize the hierarchical flow of influence from the top acquisition parameters to each predicted noise component (Fig.~\ref{fig3}k). The width of the flows corresponds to the magnitude of the SHAP values, offering a clear summary of the overall contribution of each feature. This visualization disentangles the complex causal relationships: while ISO and brightness dominate the influence on Gaussian, Salt \& Pepper, and Poisson noise, Shutter Speed and Brightness emerge as the primary drivers of Anisotropic Noise (46.10\% and 34.57\% contribution, respectively). This finding is particularly actionable for engineers, indicating that anisotropic noise (often related to motion artifacts or non-uniform processing) is best optimized not by adjusting ISO alone, but by precise optimization of exposure time and lighting conditions. The accompanying pie charts quantify these relative contributions, affirming the CoP framework’s ability to provide evidence-based, component-specific strategies for image system design and optimization.

\section{Quantitative optimization of photon-efficient protocols in in-vivo deep-tissue imaging }
\noindent 
The ultimate utility of the CoP framework is its capacity for rigorous, quantitative, multi-parametric assessment of technical interventions in high-stakes scientific imaging, where subtle signal-to-noise compromises can invalidate critical findings. We demonstrate this by applying CoP to three-photon microscopy (3PM), an advanced technique for deep-tissue \textit{in vivo} imaging often limited by severe signal attenuation and depth-dependent noise accumulation~\citep{horton2013vivo}. Specifically, we used CoP to evaluate the effectiveness of an optimized imaging strategy: utilizing a depth-gained excitation laser power to compensate for signal loss versus the conventional fixed excitation power protocol.

Qualitatively, the benefit of the intervention is immediately apparent in the 3D volume renderings of a mouse brain, showing vasculature (red) and neurons (green). The fixed excitation image stack (Fig.~\ref{fig4}a) exhibits pronounced degradation and loss of fine structure in deeper layers, whereas the depth-gained excitation stack (Fig.~\ref{fig4}b) maintains high-fidelity visibility of the finer neural and vascular structures deep into the tissue (Supplementary Fig.~\ref{figS6}).

Quantitatively, CoP's depth-resolved analysis reveals distinct noise profiles (Fig.~\ref{fig4}c--g). For Gaussian and Salt \& Pepper noise, both excitation strategies yield consistent profiles across the imaging depth, with no significant divergence observed (Fig.~\ref{fig4}c, d). In contrast, distinct depth-dependent behaviors emerge for Poisson, Quantization, and Anisotropic noise, particularly beyond $600 \,\mu\mathrm{m}$. While aggregate statistics ($100$--$800 \,\mu\mathrm{m}$) suggest moderate differences, these figures mask critical transitions occurring at the white matter interface. The abrupt divergence trends align with the imaging focal plane crossing the corpus callosum, where signal collection is hindered by photon scattering. Consequently, under Fixed Excitation, Anisotropic noise scales positively with depth, surging as the signal-to-noise ratio degrades. Crucially, the Depth-Gained Excitation strategy effectively linearizes this relationship. In the deep tissue layers ($600$--$800 \,\mu\mathrm{m}$), Gained Excitation significantly suppressed Anisotropic noise to $6.22 \pm 0.51\%$ (vs. $9.56 \pm 2.29\%$ in Fixed) and reduced Quantization noise to $2.22 \pm 0.46\%$ (vs. $3.45 \pm 0.28\%$ in Fixed). Although Poisson noise increased in this region due to signal amplification ($3.52\%$ vs. $1.69\%$), the strategy successfully mitigated the structurally complex Anisotropic noise, yielding a more stable profile across deep layers.

Crucially, CoP allows us to rigorously quantify the success of intervention in decoupling noise accumulation from the depth coordinate. We analyzed the depth-noise gradient ($\Delta$noise/µm) for both protocols (Fig.~\ref{fig4}h). While fixed power resulted in a steep positive gradient for all noise types, the gained-power protocol suppressed this slope to near-zero or negative values (P $<$ 0.001), directly demonstrating a successful intervention against depth-dependent noise. This is further validated by the $R^2$ of the linear fit: the dramatic reduction in $R^2$ from 0.42 ± 0.25 to 0.09 ± 0.08 under gained-power condition, proves that the imaging noise has been statistically de-correlated form the depth coordinate (Fig.~\ref{fig4}i).

Finally, we used Cohen’s $f^2$~\citep{chatfield1986exploratory,dosovitskiy2020image} to assess a statistically robust measure of the Intervention Effect Size, a metric for the overall practical significance of the gained-excitation compensation (Fig.~\ref{fig4}j). The effect size is classified as Large ($\geq$0.35) for both Quantization and Anisotropic noise, indicating that the intervention had its most profound impact on these specific noise mechanisms. The effect size for Gaussian noise is classified as medium-small ($<$0.35). This component-specific granularity provides actionable data for optimizing excitation protocols, hardware design, and algorithmic implementations. The CoP framework thus serves not only as a diagnostic tool but as a precision analytical platform for optimizing advanced scientific instrumentation.

\section{Discussion}
\noindent
This work fundamentally redefines the role of noise in imaging, establishing Noisomics, powered by the Contrastive Pre-trained (CoP) Foundation Model, as a critical new paradigm in computational imaging science. We have unequivocally demonstrated that noise is not merely an unavoidable impediment to clear vision but rather a high-dimensional, information-rich signature that, when properly decoded, harbors immense diagnostic and analytical potential. The CoP framework enables the accurate diagnosis of subtle hardware limitations, the precise optimization of complex acquisition parameters, and, most profoundly, the revelation of hidden scientific truths previously obscured by the very stochasticity we sought to suppress.

Our findings in smartphone photography (Fig.~\ref{fig3}) illuminate a critical, yet often overlooked, vulnerability within the burgeoning field of mobile health and point-of-care diagnostics. We have shown, with quantitative rigor, how subtle economic considerations in hardware design (e.g., compromises in sensor gain or low-cost illumination characteristics) directly translate into predictable, quantifiable noise signatures. The ability to precisely trace a specific diagnostic uncertainty back to its physical and operational origins, such as linking Anisotropic noise to shutter speed (Fig.~\ref{fig3}k), exemplifies a new era of precision system diagnostics. This capability extends beyond generic fault detection, enabling the pinpointing of exact physical and operational factors contributing to image degradation and diagnostic uncertainty in consumer devices used for clinical screening.

Furthermore, our advancements in deep-tissue three-photon microscopy optimization (Fig.~\ref{fig4}) convincingly demonstrate the proactive power of Noisomics. CoP acts as an architect, adeptly guiding system design and operational protocols beyond empirically set limits. By providing component-specific feedback, it confirmed that depth-gained excitation not only generally reduces noise but specifically and maximally combats Quantization and Anisotropic noise with a large effect size (Fig.~\ref{fig4}j). This adaptive, noise-aware guidance achieved a significant advancement in deep-tissue signal fidelity, intrinsically minimizing phototoxicity and paving the way for more delicate, prolonged, and nuanced in-vivo studies previously considered unattainable under conventional operating principles. CoP thus provides a robust, physics-informed analytical platform for optimizing advanced scientific instrumentation.

Beyond these specific demonstrations, the CoP framework provides a practical and extensible toolkit for the broader imaging community. The robust generalization proven across twelve diverse out-of-domain datasets (Fig.~\ref{fig2}a) confirms its utility as a foundational model for assessing and improving image quality across fields from medical to satellite imaging. It empowers researchers and clinicians to quantitatively assess and mitigate hidden trade-offs, such as those introduced by sophisticated black-box computational denoising algorithms that, while improving aesthetic appeal, can unintentionally obscure crucial diagnostic information.

Despite its transformative capabilities, the current CoP model, like any foundation model, operates within certain limitations that delineate clear avenues for future research. The model’s unparalleled performance is intrinsically contingent on the diversity and richness of its noise genome and training data. While CoP establishes a robust baseline for Noisomics, we acknowledge that our current noise genome, comprising five canonical types, represents a foundational rather than exhaustive set. Physical reality encompasses complex, entangled phenomena that may not map linearly to these predefined categories. However, the power of CoP lies in its learned latent representations. Even for noise sources outside the training set, the model’s contrastive objective enforces a separation from the signal manifold.

Future work will focus on expanding the noise genome and employing unsupervised clustering within the latent space to discover and categorize novel, previously undefined noise signatures, further advancing the frontier of quantitative imaging Noisomics. Expanding this dataset to include an even broader spectrum of imaging physics will undoubtedly enhance CoP’s universality and robustness. Furthermore, while the learned CoP feature space successfully disentangles noise features from content (Fig.~\ref{fig2}b, c), the complete interpretability and semantic mapping of all latent noise factors within the hyperspace remain an active and evolving challenge. While many factors clearly correspond to known noise sources, some dimensions may represent complex, entangled phenomena or higher-order statistical properties that require further refinement and dedicated domain expert analysis for full elucidation. Future work will focus on developing more sophisticated techniques for interpreting these complex latent factors and assigning precise physical or biological meanings. 

Finally, while the initial computational investment for training CoP is substantial (extensive GPUs over a period of weeks), which is a hallmark of large-scale foundation models, the resulting inference efficiency enables real-time applications such as adaptive microscopy and on-device diagnostic checks, mitigating the barrier to its widespread application. By turning noise from a source of error into a source of insight, Noisomics establishes a new computational frontier for the next generation of intelligent imaging systems.

\bibliographystyle{plainnat}
\bibliography{paper}

\newpage
\noindent
\section*{Methods}
\subsection*{The CoP foundation model and Noisomic engine}
\textbf{Model Architecture: }The neural network is based on the standard Vision Transformer (ViT-B) architecture~\citep{dosovitskiy2020image}. The model accepts input images of resolution 192×192. The model comprises a 12-layer transformer encoder with an embedding dimension of 768 and 12 attention heads. It utilizes absolute positional embeddings and a standard learnable classification token for feature aggregation. The architecture incorporates layerscale (initialization value 0.1) and stochastic depth decay rule (rate 0.1) to enhance model convergence and generalization. The final output layer projects the high-dimensional features from the classification token into 6 class probabilities.

\textbf{Expandable Noisomic Engine: }To assess the robustness of the model against diverse image degradations, we developed an Expandable Noisomic Engine (ENE). This module operates as a stochastic augmentation pipeline that sequentially applies a spectrum of noise perturbations to the input image tensor $X \in [0, 1]^{C \times H \times W}$. The engine defines a composite noise model governed by a strength vector $\eta$, which is derived from a random normal distribution via the Softmax function:
\begin{equation*}
    \eta = \operatorname{Softmax}(z), \quad \text{where } z \sim \mathcal{N}(0, I_k)
\end{equation*}
where $k=6$ represents the number of noise primitives (including the clean signal). The final corrupted image $X_{\text{final}}$ is obtained by iteratively passing the image through the sequence of noise functions $\{ f_1, f_2, \ldots, f_k \}$, where each function $f_i$ receives a specific intensity $\eta_i$. The update rule at step $i$ is defined as:
\begin{equation*}
    X_i = \operatorname{clamp}\left(f_i\left(X_{i-1}, \eta_i\right), 0, 1\right)
\end{equation*}

\noindent where $X_0$ is the clean input image, and the $\operatorname{clamp}(\cdot)$ operator ensures pixel values remain within the valid range $[0, 1]$. The specific noise primitives $f_i$ are defined as follows:

\begin{enumerate}
    \renewcommand{\labelenumi}{\textbf{\alph{enumi}.}} 

    \item \textbf{Gaussian noise:} Additive white Gaussian noise is applied with a standard deviation determined by the parameter $\eta_G$:
    \begin{equation*}
        f_{\text{gauss}}(X, \eta_G) = X + N, \quad N \sim \mathcal{N}(0, \eta_G^2)
    \end{equation*}

    \item \textbf{Salt \& Pepper noise:} Impulse noise is introduced based on a threshold $\eta_S$. Let $R \sim \mathcal{U}(0, 1)$ be a random tensor of the same shape as $X$. The transformation is defined as:
    \begin{equation*}
        f_{\text{sp}}(X, \eta_S)_{ijk} = 
        \begin{cases} 
            0, & \text{if } R_{ijk} < \eta_S \\
            1, & \text{if } R_{ijk} > 1-\eta_S \\
            X_{ijk}, & \text{otherwise}
        \end{cases}
    \end{equation*}

    \item \textbf{Poisson noise:} We implemented an additive signal-dependent noise model. Scaled by a parameter $\eta_P$, the noise is generated from a Poisson distribution and added to the original signal:
    \begin{equation*}
        f_{\text{poisson}}(X, \eta_P) = X + \operatorname{Poisson}(\eta_P \cdot X)
    \end{equation*}

    \item \textbf{Quantization noise:} To simulate low-bitrate artifacts, we applied uniform noise followed by a quantization step with parameter $\eta_Q$:
    \begin{equation*}
        f_{\text{quant}}(X, \eta_Q) = \eta_Q \cdot \left(\frac{X}{\eta_Q} + U\right), \quad U \sim \mathcal{U}(0, 1)
    \end{equation*}

    \item \textbf{Anisotropic noise:} We simulated spatially correlated noise by convolving Gaussian noise with a fixed $5 \times 5$ Gaussian smoothing kernel $K$. The kernel approximates a 2D Gaussian distribution (normalized such that $\sum K=1$). The noise is scaled by $\eta_A$:
    \begin{equation*}
        f_{\text{aniso}}(X, \eta_A) = X + (N' \ast K), \quad N' \sim \mathcal{N}(0, \eta_A^2)
    \end{equation*}
    where $\ast$ denotes the convolution operation.

    \item \textbf{Clean signal:} A pass-through operation representing the absence of noise for that component.
\end{enumerate}

This sequential application ensures that the generated samples cover a complex manifold of degradations, ranging from single-source noise to mixed-noise scenarios, governed by the stochastic intensity vector $\lambda$. Different parameter combinations and visualizations can be found in Supplementary Figs.~\ref{figS1}--\ref{figS2}. It is worth noting that these noise components are not fixed and can be expanded according to the actual situation.

\subsection*{Training and evaluating details}
\textbf{Training Data Corpus: }The foundation model was trained on a subset of the ImageNet-1k~\citep{deng2009imagenet} dataset, serving as the source of clean semantic signals. The ENE generated over 1 million unique noisy instances by sampling from the Noisomic parameters (server as the ground truth for supervised fine-tuning). To benchmark data efficiency, we also trained models on subsets ranging from 100 to 100,000 samples (Fig.~\ref{fig1}c).

\textbf{Two-Stage Training Protocol: }The training proceeds in two distinct phases:
\begin{enumerate}
    \renewcommand{\labelenumi}{\textbf{\alph{enumi}.}} 
    \item Unsupervised Contrastive Pre-training: The encoder is pre-trained on the synthetic dataset generated by the ENE. The model optimizes a contrastive objective to cluster embeddings based on noise type and magnitude, effectively pushing noise features separated to the semantic manifold. The network was optimized to minimize a modified InfoNCE~\citep{oord2019representation} loss as follows:
    \begin{equation*}
        \mathcal{L}_{\text{contrast}} = -\frac{1}{N} \sum_{i=1}^{N} \log \frac{\exp(\mathbf{z}_{A+N_A}^{(i)} \cdot \mathbf{z}_{B+N_A}^{(i)} / \tau)}{\exp(\mathbf{z}_{A+N_A}^{(i)} \cdot \mathbf{z}_{B+N_A}^{(i)} / \tau) + \sum_{j=1, j \neq i}^{N} \exp(\mathbf{z}_{A+N_A}^{(i)} \cdot \mathbf{z}_{A+N_B}^{(j)} / \tau)}
    \end{equation*}
    \noindent here $\tau=0.1$ denotes the temperature scaling factor, dot ($\cdot$) represents the dot product, and the denominator aggregates similarities against the contaminated samples in the batch. This formulation explicitly disentangles noise artifacts from the image representation.

    \item Supervised Fine-tuning: The encoder weights are frozen to preserve the learned noise-manifold topology. The Parametric Quantification Head is attached and trained using supervised learning with Mean Squared Error (MSE) loss to regress the precise ground-truth parameters of the noise vectors applied during synthesis.
    \begin{equation*}
        \mathcal{L}_{\text{noise}} = \frac{1}{N} \sum_{i=1}^{N} \| \mathbf{y}_i - \hat{\mathbf{y}}_i \|_2^2
    \end{equation*}
    \noindent where $\mathbf{y}_i$ represents the ground-truth noise parameter vector and $\hat{\mathbf{y}}_i$ denotes the predicted values.
\end{enumerate}

\textbf{Computational Resources: }All models were trained and evaluated on NVIDIA H20 GPUs Cluster, and the maximum peak number of GPU calls is 128. The framework was implemented in PyTorch~\citep{paszke2019pytorch}. The large-scale pre-training (1M+ samples) required approximately 1,500 GPU hours, while inference operates in real-time.

\subsection*{Generalization verification and analysis}
\textbf{Out-of-Domain Generalization Datasets: }To validate device-agnostic performance, we curated 12 diverse Out-of-Domain (OOD) datasets encompassing medical, scientific, and artistic modalities: Chest X-Ray, Histopathology, Retina fundus, Alzheimer’s MRI, Microscopy, Universe, Satellite imagery, Describable Textures, Sketch Scene, Manga, Underwater, and Natural Scene. None of these datasets were seen during training. Detailed OOD dataset information can be found in Supplementary Table~\ref{tabS1}.

\textbf{Feature Space Analysis: }We employed t-Distributed Stochastic Neighbor Embedding (t-SNE) to visualize the high-dimensional latent space in 2D. Feature distribution alignment was quantified using the Maximum Mean Discrepancy (MMD) metric with a radial basis function (RBF) kernel, calculating the distance between the mean embeddings of different OOD datasets to assess domain invariance.

\textbf{Evaluation and Analysis Metrics: }To rigorously assess the performance of the noise estimation module, we evaluated the model using four statistical metrics: Root Mean Squared Error (RMSE), and the Coefficient of Determination ($R^2$). The $R^2$ score specifically indicates the proportion of variance in the noise parameters predictable from the input image features. To verify generalization robustness, this evaluation was extended to 12 heterogeneous OOD datasets with performance differences validated via paired two-sided t-tests (P<0.001). Beyond scalar metrics, we scrutinized the latent space topology using t-Distributed Stochastic Neighbor Embedding (t-SNE) for semantic visualization and computed the Maximum Mean Discrepancy (MMD) to quantify domain invariance and the mitigation of distributional shifts. Furthermore, fine-grained prediction fidelity was analyzed through Kernel Density Estimation (KDE) of the joint output distribution, complemented by fitting a Gaussian probability density function to the prediction residuals, utilizing the derived mean bias ($\mu$) and standard deviation ($\sigma$) as quantitative indicators of estimation stability and precision.

\subsection*{Smartphone Photography Data Protocol and Experiment}
\textbf{Dataset: }To validate the model’s applicability in real-world mobile photography scenarios, we utilized the SIDD Medium benchmark dataset~\citep{abdelhamed2018highquality}, which comprises 320 pairs of noisy and ground-truth images captured by varying smartphone devices.

\textbf{Sampling and Preprocessing Strategy: }To ensure computationally efficient yet statistically robust noise estimation, we implemented a stochastic windowing protocol. For every noise estimation experiment, instead of processing the full-resolution image, we randomly cropped five independent windows of resolution 192×192 pixels from the original raw data. The noise parameters were computed individually for each crop, and their arithmetic mean was adopted as the final image-level estimate.

\textbf{Experimental Subsets: }For the no-reference image quality assessment analysis (Fig.~\ref{fig3}a–d), we curated a specific subset to investigate sensitivity-dependent performance. We filtered the dataset for samples with complete metadata and sufficient representation, resulting in 184 image pairs (112 captured at ISO 100 and 72 at ISO 800). For all other quantitative evaluations and interpretability analyses, the complete dataset of 320 pairs was utilized to maximize the diversity of acquisition conditions.

\textbf{Interpretability and Feature Attribution Analysis: }To disentangle the complex relationships between hardware acquisition parameters and the model-predicted noise profiles, we employed a three-tiered analysis framework.
\begin{enumerate}
    \renewcommand{\labelenumi}{\textbf{\alph{enumi}.}}
    
    \item \textbf{Correlation Analysis:} We quantified the linear dependencies between acquisition metadata (ISO, shutter speed, brightness, smartphone type, and temperature) and the estimated levels of the five noise components. These relationships were visualized via a correlation heatmap (Fig.~\ref{fig3}e) to identify dominant covariates and assess the strength of associations between capture settings and noise magnitude.

    \item \textbf{SHAP Feature Attribution:} To provide a granular explanation of the model’s decision-making process, we utilized SHAP. Summary plots (Fig.~\ref{fig3}f--j) were generated to compute the marginal contribution and directionality of each acquisition parameter. This approach allows for the assessment of how specific feature values drive or suppress the prediction of distinct noise types.

    \item \textbf{Global Influence Visualization (Sankey Diagram):} To illustrate the macroscopic flow of influence, we constructed a Sankey diagram (Fig.~\ref{fig3}k). In this visualization, the width of the connecting flows is proportional to the mean absolute SHAP value of each feature. This metric was employed to quantify and rank the relative importance of each acquisition parameter in the overall mechanism of image degradation.
\end{enumerate}

\subsection*{Deep-Tissue Imaging Using Three-Photon Microscopy}
\textbf{Experimental Setup: }Three-photon \textit{in vivo} volumetric imaging was performed using a commercial multi-photon microscope system (DeepVision, MicroLux, China). The optical architecture was designed to achieve a 1.6 mm × 1.6 mm field of view (FOV) using a 16× water-immersion objective (CFI75 LWD 16X W, Nikon), with imaging performed at 0.5 frames per second. To optimize signal-to-noise ratios in deep tissue, the detection path employed a near-axial, high-efficiency design with a collection angle of up to 15 degrees, ensuring the maximal capture of both ballistic and scattered fluorescence photons via high-sensitivity GaAsP photomultiplier tubes (H16201P-40, Hamamatsu). Three-photon imaging employed a 1,300 nm femtosecond pulsed laser (Cronus3P, Light Conversion) with 50-fs pulse duration and 1-MHz repetition rate. To simultaneously acquire green fluorescent protein (GFP) and vasculature (Texas Red) signals, emitted photons were spectrally separated into green (510±30) and red (605±35 nm) channels, respectively. Volumetric imaging was performed over a depth range of 0–1300 µm with a 4 µm inter-plane interval, using a motorized objective stage. Each plane was captured at a 512×512-pixel resolution. Laser power was dynamically adjusted (5–60 mW) to compensate for signal attenuation at increasing depths.

\textbf{Animal preparation and labeling: }Experimental subjects consisted of wild-type C57BL6/J mice (8–10 weeks postnatal) maintained under controlled environmental conditions with a 12:12 h light-dark cycle and ad libitum access to food and water. For three-photon functional imaging, mice received stereotaxic injections of pAAV-hSyn-jGCaMP7s-WPRE. The viral vector (200 nL per site) was delivered via glass micropipettes (RWD Life Science, China) into the target coordinates: AP -2.0 mm, ML -1.76 mm, at depths (DV) of -0.4, -0.8, -1.2, and -1.6 mm. After a 2–3 week recovery period to allow for viral expression, the skull was exposed and a cranial drill was used to remove a portion of the skull. A custom-made coverslip (~4 mm in diameter) was secured over the cranial window using dental cement. Postoperative analgesia and monitoring are provided. After surgery, mice were housed in a separate cage for 2 weeks of postoperative recovery. To enable simultaneous vasculature visualization, animals received a tail vein injection of 0.1 ml Texas Red (25 mg/ml, dextran conjugate, 70 kDa, Thermo Fisher Scientific, USA) prior to imaging sessions. All experimental procedures were approved by the Animal Welfare and Ethics Committee of the Department of Experimental Animal Science, Fudan University (Ethics Approval Number: 2024-FAET-005).

\textbf{Laser excitation protocols: }Imaging was conducted using two distinct power modulation strategies to evaluate signal penetration. In the Fixed Excitation mode, laser power was held constant ($4.3\,\mathrm{mW}$) at the initial surface level across all imaging depths. Conversely, to address signal loss in deep tissue, the Depth-Gained Excitation mode was employed. In this regime, laser power was modulated to exponentially increase with depth ($z$) to compensate for scattering-induced attenuation. This modulation followed a standard exponential energy growth curve based on the Beer-Lambert law $P(z) = P_0 \cdot e^{z/L_e}$, capped at a pre-defined safety threshold ($< 100\,\mathrm{mW}$) to prevent thermal damage to the biological tissue.

\textbf{Quantitative Analysis: }To systematically evaluate the efficacy of depth-gained excitation in mitigating scattering-induced signal degradation, we performed a comparative quantitative analysis against the fixed excitation baseline using CoP predictions across the volumetric data. We modeled the depth-dependent noise accumulation via linear regression ($y = \beta z + \alpha$), utilizing the slope coefficient ($\beta$) and Coefficient of Determination ($R^2$) to quantify the rate of noise increase and the strength of the depth-noise correlation, respectively. Furthermore, the practical impact of the intervention was rigorously stratified using Cohen's $f^2$ effect size (classified as $\text{Small} < 0.15$, $\text{Medium} < 0.35$, or $\text{Large} \ge 0.35$), with the statistical significance of regression slope differences validated using t-tests ($P < 0.001$).

\section*{Data and code availability}
The pre-trained CoP model weights, the Noise Engine source code, and the specific inference scripts used for the figures are available at https://github.com/FDU-donglab/CoP. The diverse OOD datasets used for validation are publicly available from their respective repositories (see Supplementary Table~\ref{tabS1}). The smartphone data is available at (https://abdokamel.github.io/sidd/). The three-photon microscopy raw data are available from the corresponding authors upon reasonable request.

\section*{Acknowledgements}
This work was supported by Damo Academy through Damo Academy Innovative Research Program. We thank Yao Wu from MicroLux (Shanghai) Intelligent Science \& Technology Co., Ltd. for technical assistance with the three-photon microscopy experiments.

\section*{Author contributions}
Y.G., Y.W. and C.Y. contributed to the conceptualization, coding, comparisons, and visualization. A.X. contributed to the experiments and performed data collection. F.W., Z.L. and B.D. supervised the project, contributed to the conceptualization, and designed experiments. Y.G., Y.W., C.Y., Z.L. and B.D. wrote the manuscript. All authors have read and approved the manuscript.

\section*{Funding}
This work was supported by the National Key R\&D Program of China (2022YFF0708700), the Shanghai Basic Research Special Zone Program (22TQ020), the Medical Engineering Fund of Fudan University (yg2025-key-04), the National Natural Science Foundation of China (625B2053, 62575154), the Young Scientists Fund of Beijing Natural Science Foundation (4254114).

\section*{Competing interests}
C.Y. and F.W. are employees of DAMO Academy, Alibaba Group. B.D. is a founder and equity holder of MicroLux (Shanghai) Intelligent Science \& Technology Co., Ltd. and Lishi Intelligent Science \& Technology (Shanghai) Co., Ltd. The remaining authors declare no competing interests.

% \newpage
\section*{Figures}
% ==================== Figure 1 ====================
\begin{center}
    \includegraphics[width=1\textwidth]{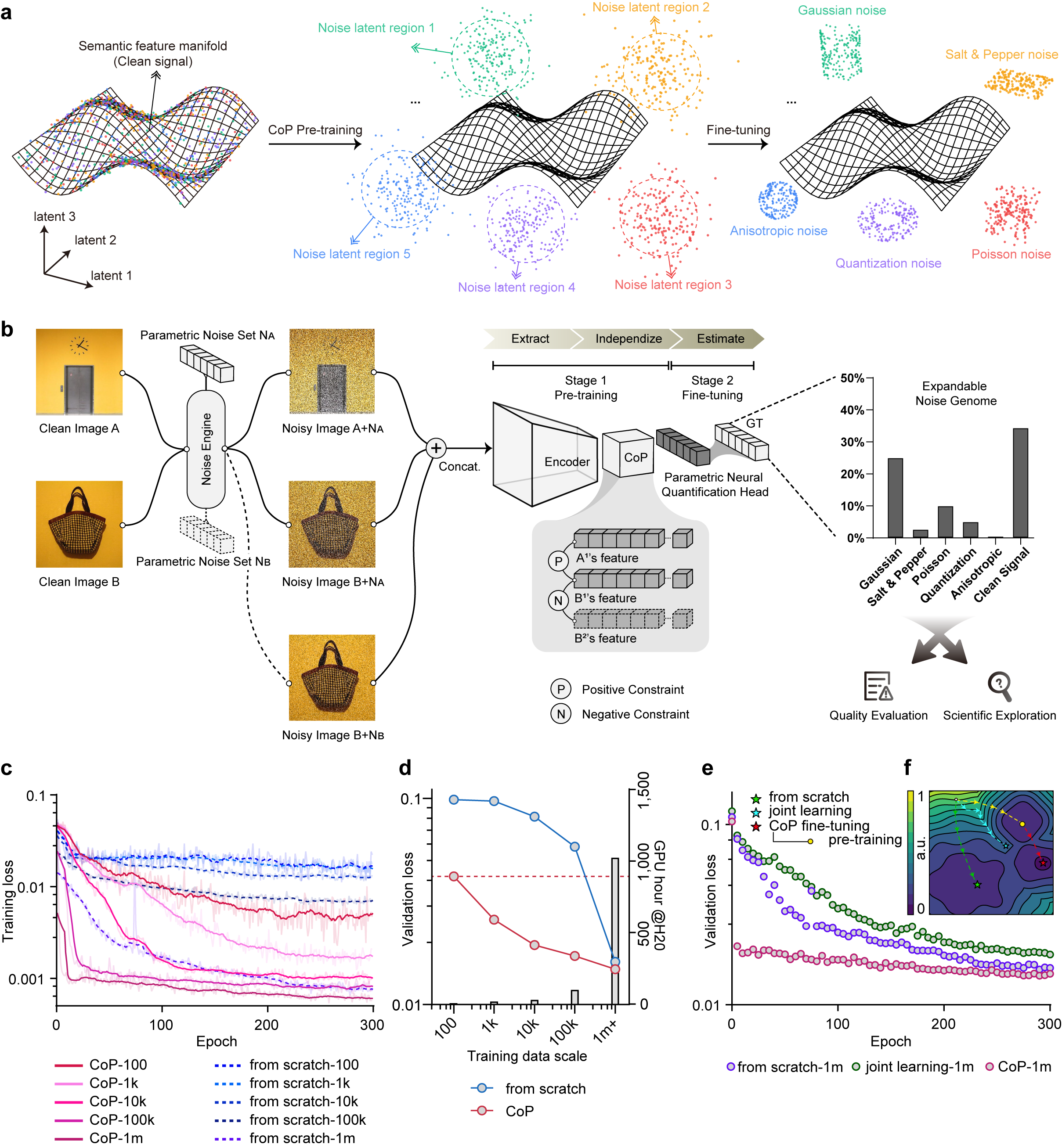}
\end{center}
\vspace{0.2em}
\captionof{figure}{\textbf{The Contrastive Pre-trained (CoP) framework for Noisomics.}} 
\label{fig1} 
\vspace{0.2em} \noindent
\textbf{a,} Conceptual visualization of the feature space evolution. The clean signal resides on a semantic feature manifold. CoP pre-training disentangles noise features from the clean signal, mapping them into distinct noise latent regions. Subsequent fine-tuning further separates and clusters these regions into specific noise categories. 
\textbf{b,} Schematic of the CoP framework. Clean images ($A, B$) are corrupted by a noise engine using parametric noise sets ($N_A, N_B$) drawn from an expandable noise genome. The resulting noisy image pairs are used in a two-stage training process. Stage 1 involves unsupervised pre-training using contrastive learning to distinguish between features derived from the same underlying noise (Positive Constraint, $P$) and those from different noise (Negative Constraint, $N$). Stage 2 fine-tunes a parametric neural quantification head to estimate the specific noise parameters. The noise genome, illustrating the distribution of noise types, facilitates applications in quality evaluation and scientific exploration. 
\textbf{c,} Training loss as a function of epoch for models trained with CoP (solid lines) versus models trained from scratch (dashed lines) across datasets of varying sizes (from 100 to 1 million samples). The CoP approach demonstrates consistently lower loss and faster convergence across all data scales. The model trained using CoP even with only 100 samples for training outperforms the conventional method that uses 100k samples for training, breaking the scaling law of noise estimation. 
\textbf{d,} Validation loss (left axis) and required GPU hours (right axis, on H20 hardware) plotted against training data scale. CoP achieves a lower validation loss than training from scratch across all data scales while also being more computationally efficient, a benefit that reduces with scaling up dataset size. 
\textbf{e,} Comparison of validation loss versus epoch for three different training strategies on the 1 million sample dataset: training from scratch, joint learning, and CoP fine-tuning. The CoP fine-tuning strategy achieves the lowest validation loss, outperforming the other methods. 
\textbf{f,} A conceptual visualization of the loss landscape showing the optimization trajectories for the training strategies presented in \textbf{d}. The trajectory of the CoP fine-tuning approach (red star) converges to a superior local minimum with a lower loss value.

% ==================== Figure 2 ====================
\newpage
\begin{center}
    \includegraphics[width=1\textwidth]{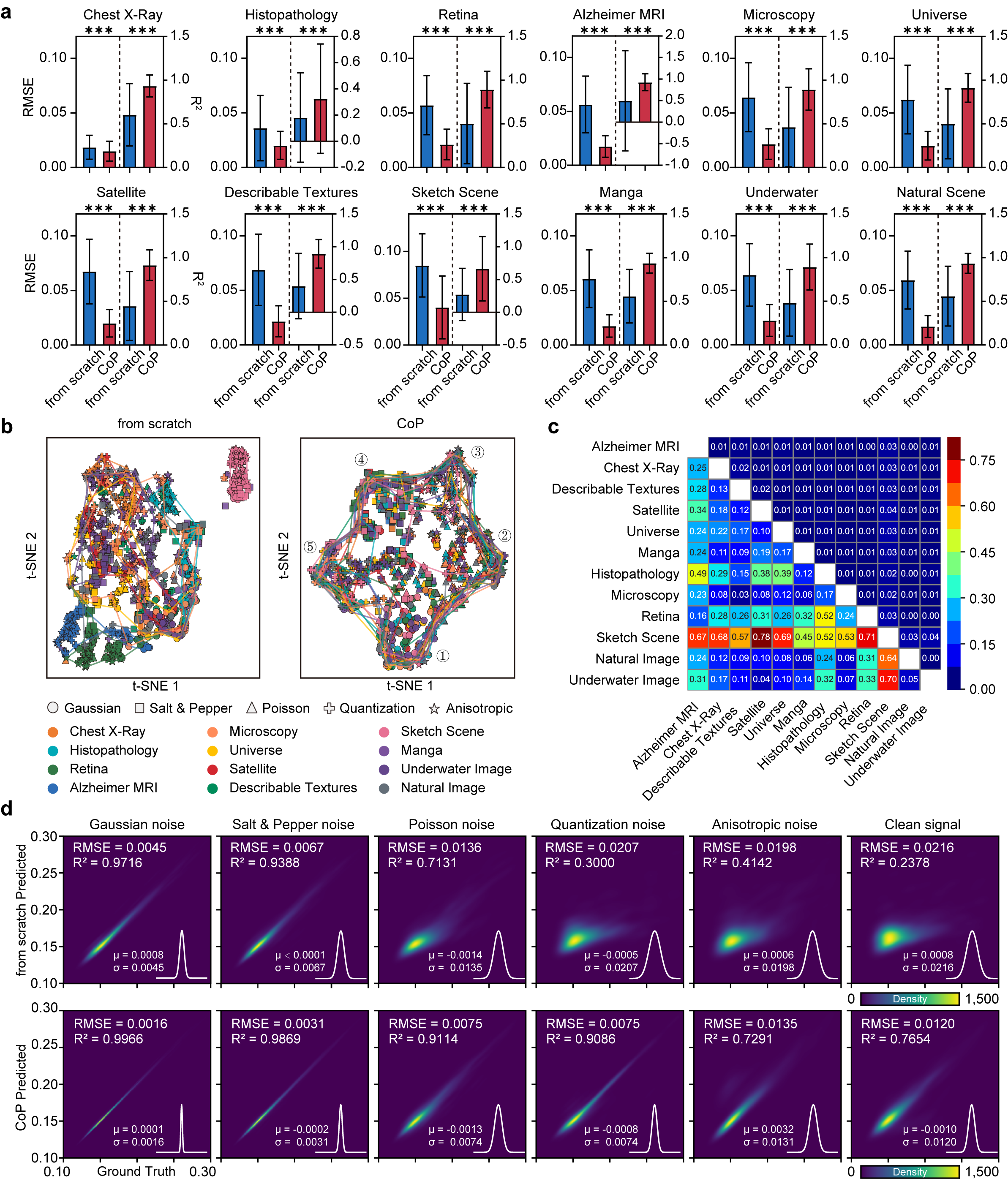}
\end{center}
\vspace{0.2em}
\captionof{figure}{\textbf{The Contrastive Pre-trained (CoP) framework for Noisomics.}} 
\label{fig2} 
\vspace{0.2em} \noindent
\textbf{a,} Performance comparison between models trained from scratch and with CoP on 12 diverse, out-of-domain image datasets. For each dataset, bar charts show the root-mean-square error (RMSE, left axis, lower is better) and the coefficient of determination ($R^2$, right axis, higher is better). CoP consistently achieves significantly lower RMSE and higher $R^2$ across all datasets, demonstrating its superior generalization ability. Error bars represent standard deviation. ***$P < 0.001$, paired t-test.
\textbf{b,} t-SNE visualizations of the feature embeddings learned by a model trained from scratch (left) versus a model trained with CoP (right). Each point represents an image, shaped by different types of dominant noise, and colored by its respective dataset. The CoP model learns a more structured and semantically organized feature space, where 5 clusters corresponding to 5 noise categories.
\textbf{c,} A similarity matrix showing the Maximum Mean Discrepancy (MMD) between the average feature vectors of different datasets, as learned by the CoP encoder. The lower left triangle represents the MMD values of the encoder pairs trained from scratch for each pair of different datasets, while the upper right triangle shows the MMD values of the encoder pairs trained with CoP for each pair of different datasets. The matrix reveals that, compared to the encoder trained from scratch, CoP has eliminated the differences between datasets, achieved excellent independence in representing noise features, and demonstrated strong generalization interpretability.
\textbf{d,} Kernel density estimation comparing the ground truth values of various noise and signal parameters against the values predicted by models trained from scratch (top row) and with CoP (bottom row). For every tested noise type and for the clean signal, the CoP model yields predictions that are more accurate and highly correlated with the ground truth, as indicated by substantially higher $R^2$, lower RMSE, and a tighter distribution of prediction errors (Gaussian fitting, $\mu$ and $\sigma$ are the mean and standard deviation of the Gaussian parameters).

% ==================== Figure 3 ====================
\newpage
\begin{center}
    \includegraphics[width=1\textwidth]{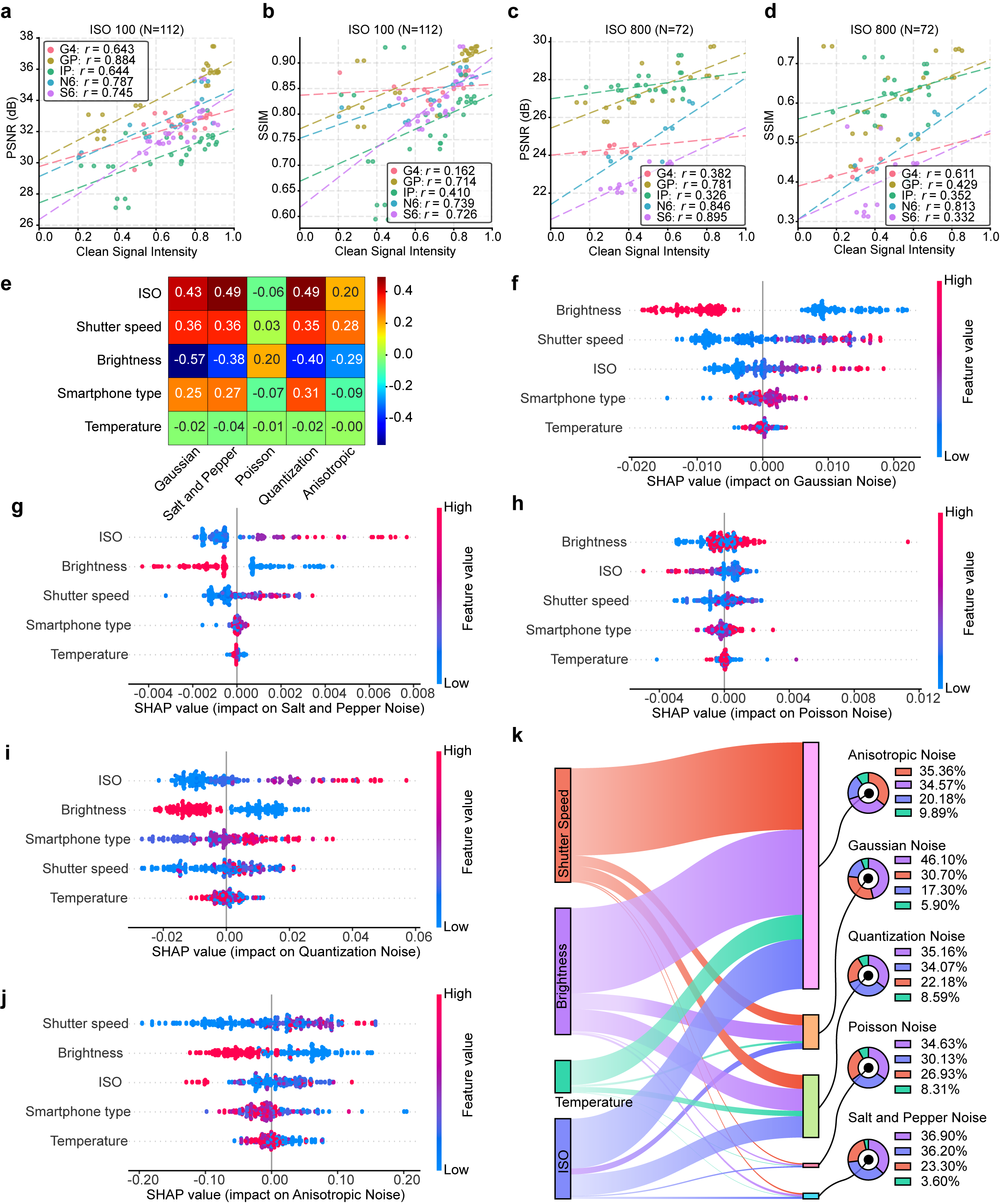}
\end{center}
\vspace{0.2em}
\captionof{figure}{\textbf{Quality evaluation of the relationships between image acquisition parameters and noise components in smartphone photography.}} 
\label{fig3} 
\vspace{0.2em} \noindent
\textbf{a--d,} Validation of the model's ``clean signal'' estimation. Scatter plots show a significant positive correlation between the estimated clean signal quality and established image quality metrics, Peak Signal-to-Noise Ratio (PSNR) and Structural Similarity Index Measure (SSIM), across real-world images taken with different smartphones ($n=112$ for ISO 100, $n=72$ for ISO 800). Dashed lines represent linear fits.
\textbf{e,} Correlation heatmap illustrating the quantitative relationships between key acquisition parameters (ISO, shutter speed, brightness, smartphone type, temperature) and the estimated levels of five different noise types. A strong positive correlation is observed between ISO and most noise components, while brightness is generally negatively correlated.
\textbf{f--j,} SHapley Additive exPlanations (SHAP) summary plots providing detailed feature attributions for the model's prediction of each noise type. These plots reveal the impact and directionality of each acquisition parameter.
\textbf{k,} A Sankey diagram visualizing the flow of influence from acquisition parameters to the predicted noise components. The width of the flows represents the magnitude of the SHAP values, effectively summarizing the overall contribution of each feature. The diagram disentangles the complex interplay of factors, showing for example that while ISO is a dominant contributor to Gaussian and Salt \& Pepper noise, shutter speed and brightness are the primary drivers of Anisotropic noise. Pie charts quantify the relative contribution of the top four features to each noise type.

% ==================== Figure 4 ====================
\newpage
\begin{center}
    \includegraphics[width=1\textwidth]{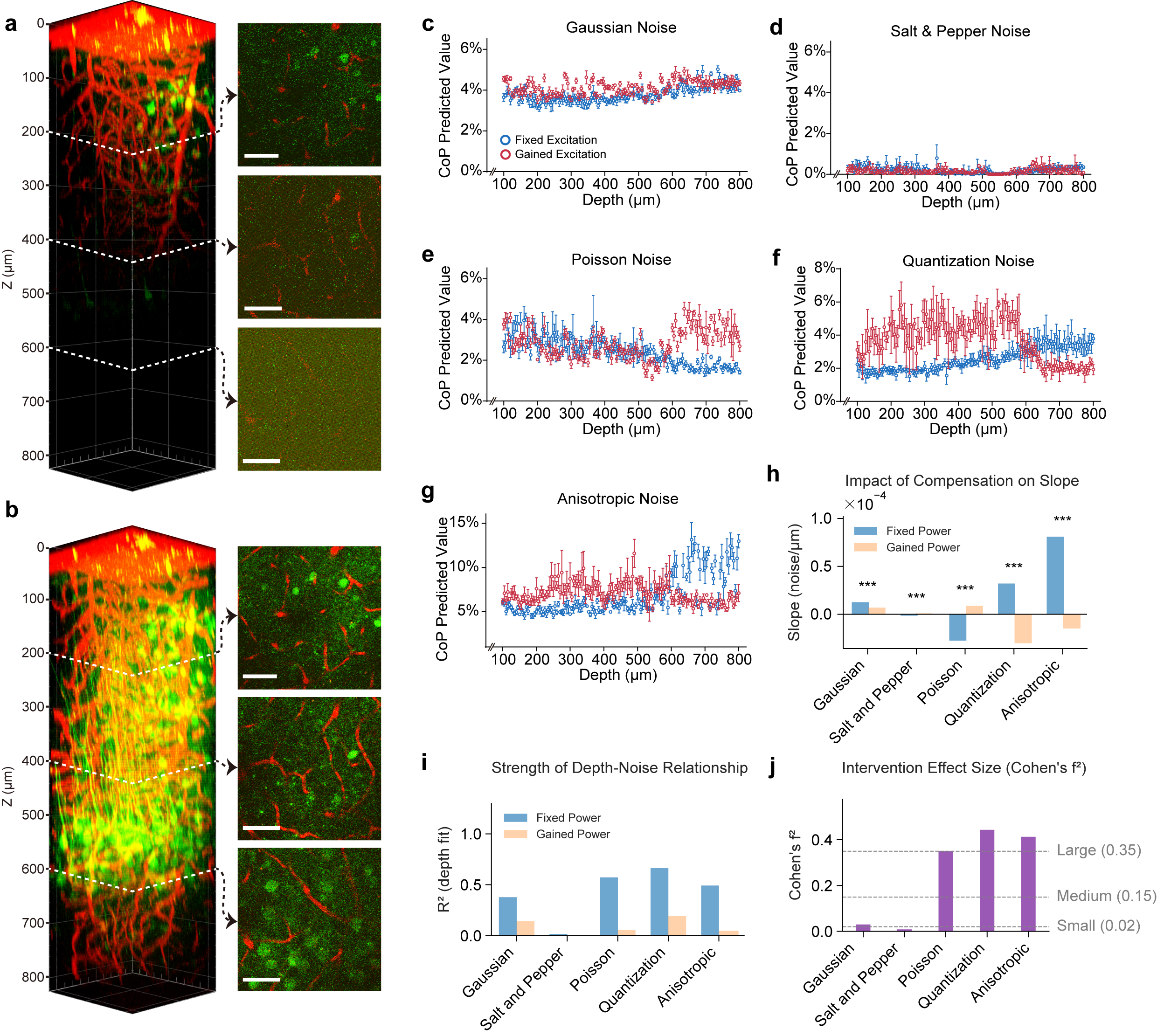}
\end{center}
\vspace{0.2em}

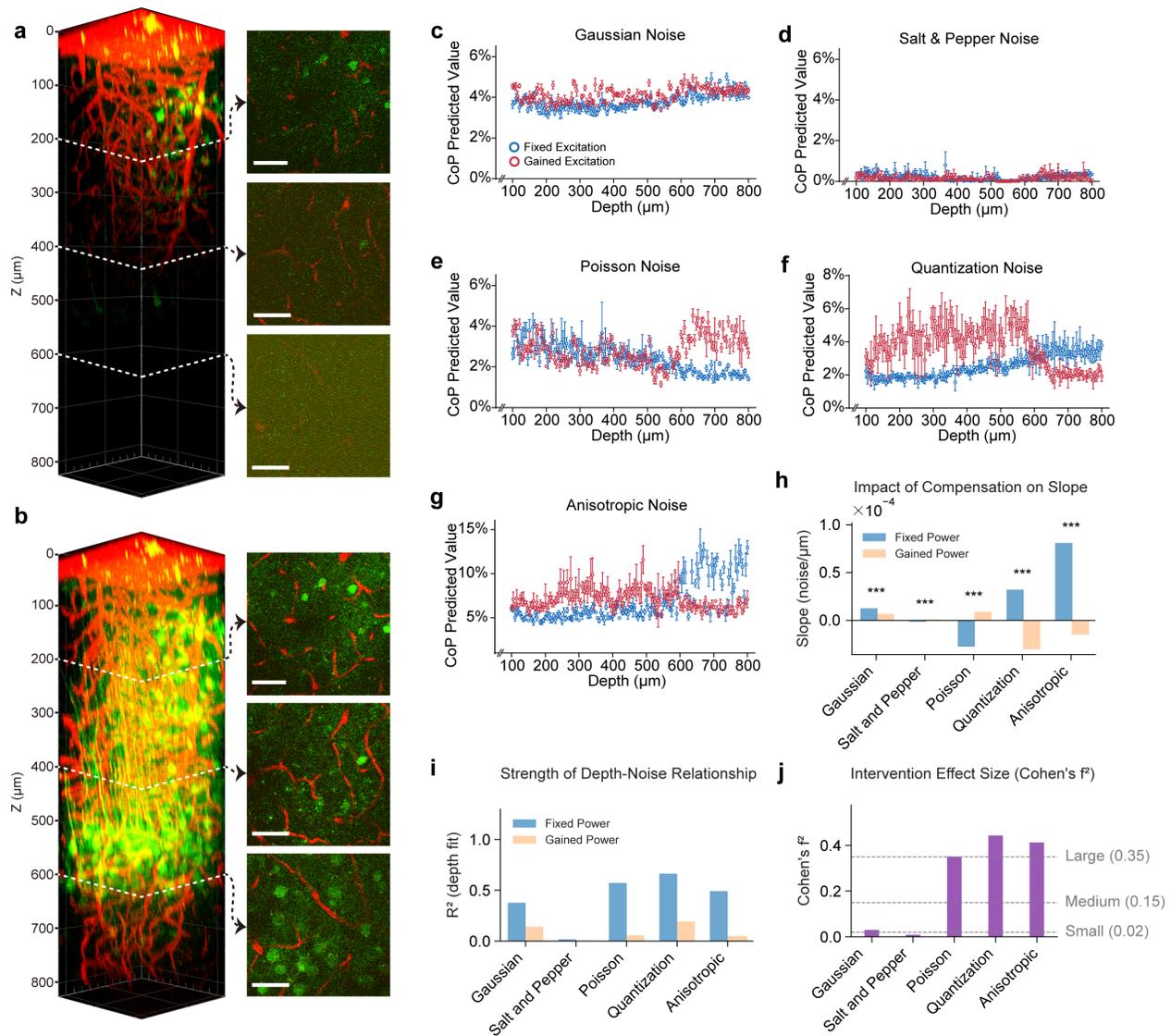
\captionof{figure}{\textbf{Quantitative analysis of noise reduction by depth-gained excitation in deep-tissue three-photon microscopy using the CoP framework.}} 
\label{fig4} 
\vspace{0.2em} \noindent
\textbf{a, b,} Three-dimensional volume renderings of three-photon microscopy image stacks acquired in a mouse brain, showing vasculature (red) and neurons (green). The images were acquired using a fixed excitation laser power (\textbf{a}) and a depth-gained excitation power (\textbf{b}). Insets display representative $x$--$y$ planes at different depths, illustrating the improved signal preservation and reduced noise in the deeper tissue achieved with gained excitation (\textbf{b}). Scale bars, $100 \,\mu\mathrm{m}$.
\textbf{c--g,} Depth-resolved quantification of five different noise components predicted by the CoP model, comparing fixed excitation (blue) and gained excitation (red) strategies. The plots show the CoP Predicted Value (\%) as a function of depth ($\mu\mathrm{m}$) for Gaussian noise (\textbf{c}), Salt \& Pepper noise (\textbf{d}), Poisson noise (\textbf{e}), Quantization noise (\textbf{f}), and Anisotropic noise (\textbf{g}). Gained excitation (red) generally yields a lower or less depth-dependent CoP Predicted Value compared to fixed excitation (blue), notably mitigating the increase in Gaussian (\textbf{c}), Quantization (\textbf{f}), and Anisotropic (\textbf{g}) noise with imaging depth. Data are presented as mean $\pm$ SD.
\textbf{h,} Gained power significantly reduces the positive slope for all noise types, often to near-zero or negative values, indicating a successful decoupling of noise accumulation from imaging depth. Statistical significance is indicated by asterisks (***$P < 0.001$).
\textbf{i,} Strength of Depth-Noise Relationship. Bar chart showing the $R^2$ from a linear fit of the predicted noise value as a function of depth for fixed power (blue) and gained power (orange). The substantial reduction in $R^2$ for gained power demonstrates the intervention's effectiveness in minimizing the depth-dependent nature of the noise.
\textbf{j,} Intervention Effect Size (Cohen's $f^2$). Bar chart quantifying the overall effect size of the depth-gained excitation intervention on the five noise components. The effect is classified as Large ($>0.35$) for Quantization and Anisotropic noise, and Medium to Small for the other components, highlighting the differential impact of the compensation strategy.

\newpage
\setcounter{figure}{0}
\renewcommand{\thefigure}{\arabic{figure}}
\renewcommand{\theHfigure}{S\arabic{figure}}
\captionsetup[figure]{name={Supplementary Figure}}
\setcounter{table}{0}
\renewcommand{\thetable}{\arabic{table}}
\renewcommand{\theHtable}{S\arabic{table}}
\captionsetup[table]{name={Supplementary Table}}
\section*{Supplementary Information}
% ==================== Figure S1 ====================
\begin{center}
    \includegraphics[width=1\textwidth]{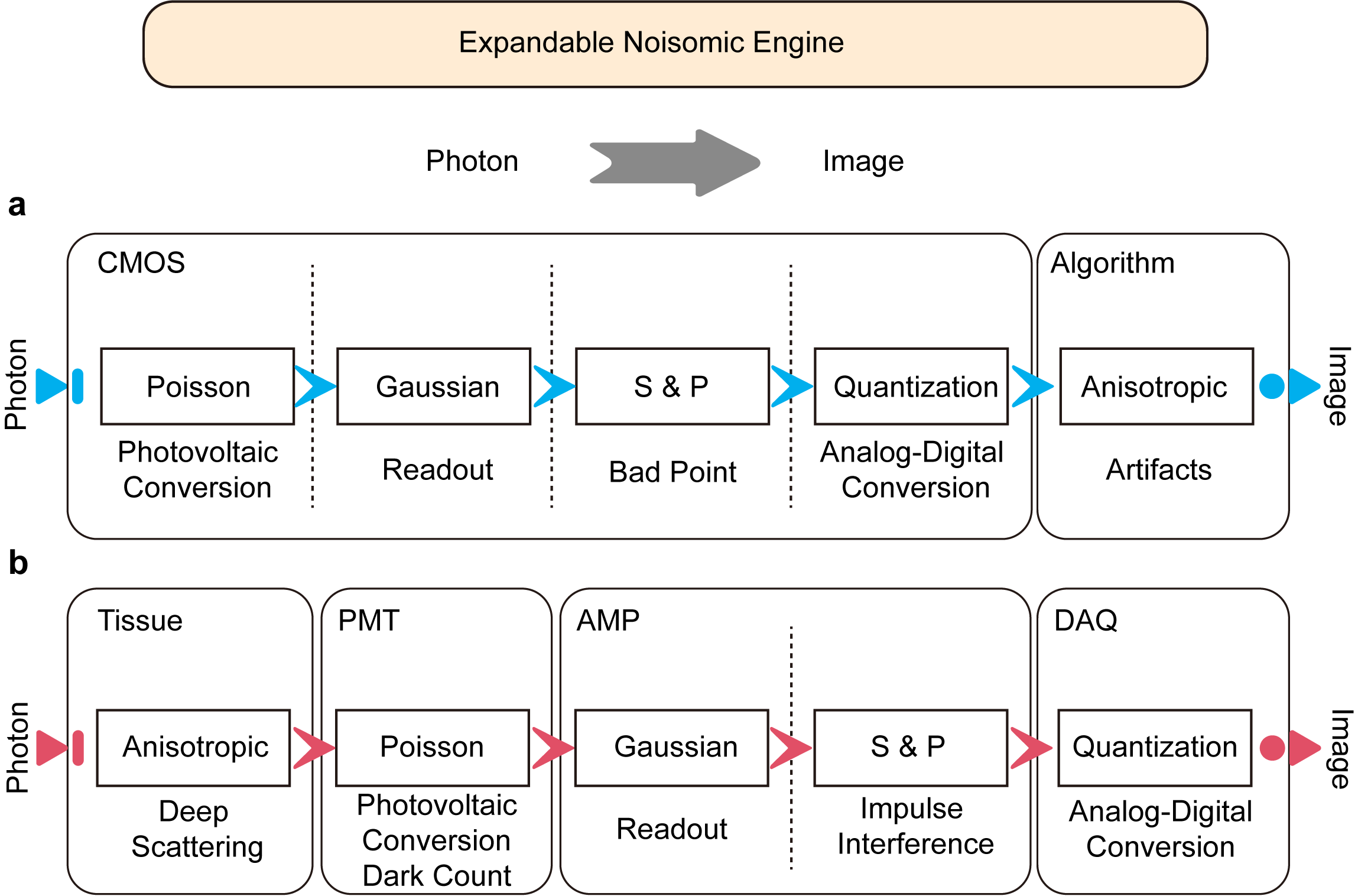}
\end{center}
\vspace{0.2em}
\captionof{figure}{\textbf{Schematic illustration of the Expandable Noisomic Engine (ENE).}} 
\label{figS1} 
\vspace{0.2em} \noindent
{The diagram depicts the physics-based noise simulation pipelines for two distinct imaging scenarios modeled by ENE. \textbf{a,} The noise generation process for standard CMOS sensors, sequentially simulating noise arising from photovoltaic conversion (Poisson), circuit readout (Gaussian), sensor defects (Salt \& Pepper), analog-to-digital conversion (Quantization), and algorithmic artifacts (Anisotropic). \textbf{b,} The pipeline for deep tissue imaging systems (e.g., PMT-based), which explicitly models photon scattering in tissue (Anisotropic), detector shot noise (Poisson), amplifier thermal noise (Gaussian), impulse interference (Salt \& Pepper), and digitization errors (Quantization).}

% ==================== Figure S2 ====================
\newpage
\begin{center}
    \includegraphics[width=0.75\textwidth]{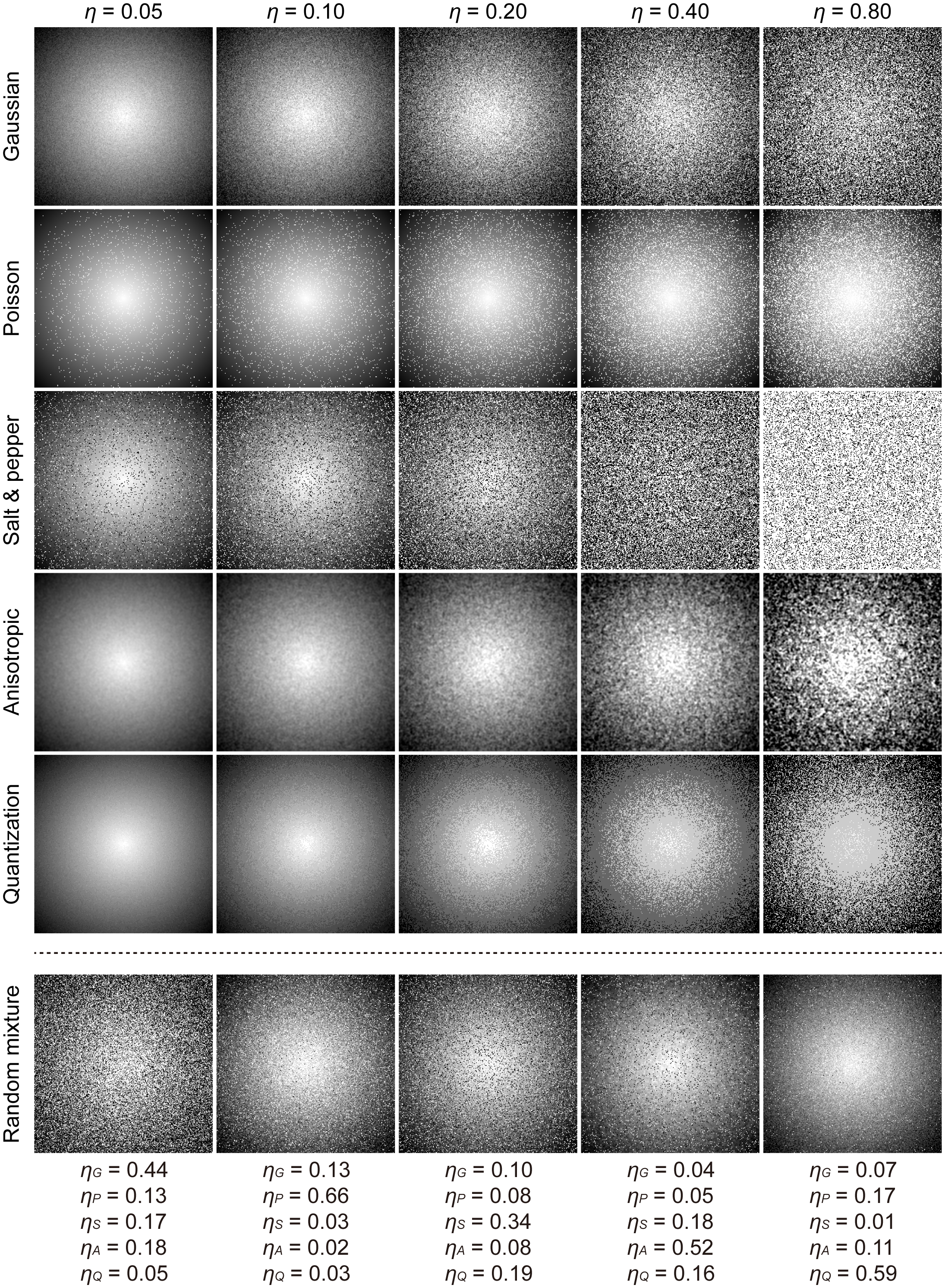}
\end{center}
\vspace{0.2em}
\captionof{figure}{\textbf{Visualization of the noise generated by ENE.}} 
\label{figS2} 
\vspace{0.2em} \noindent
{Visual examples of the five canonical single-source noise primitives modeled by the Expandable Noisomic Engine (ENE): Gaussian, Poisson, Salt \& Pepper, Anisotropic, and Quantization noise. The columns from left to right display increasing noise intensity levels ($\eta$), ranging from 0.05 to 0.80, applied to a standard clean signal. Examples of composite noise mixtures generated by the ENE, illustrating the model's capability to simulate complex, multi-parametric degradation scenarios. The specific parameter values for each noise component (Gaussian $\eta_G$, Poisson $\eta_P$, Salt \& Pepper $\eta_S$, Anisotropic $\eta_A$, and Quantization $\eta_Q$) used to generate the mixture are listed below each corresponding image. $\alpha$ represents the scaling factor.}

% ==================== Figure S3 ====================
\newpage
\begin{center}
    \includegraphics[width=0.75\textwidth]{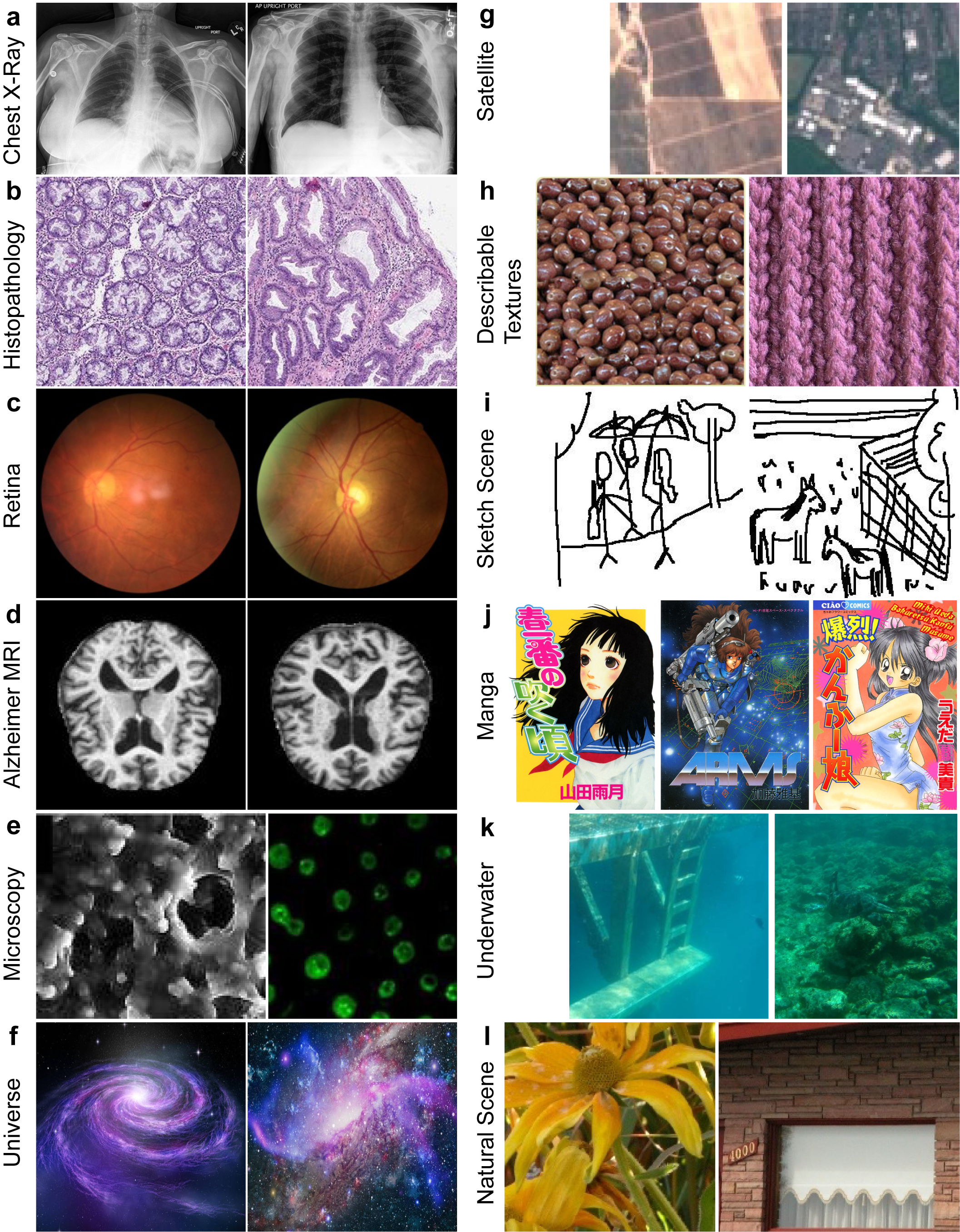}
\end{center}
\vspace{0.2em}
\captionof{figure}{\textbf{Representative samples from the twelve OOD datasets used for evaluation.}} 
\label{figS3} 
\vspace{0.2em} \noindent
{The figure displays example images from diverse domains: \textbf{a,} Chest X-Ray; \textbf{b,} Histopathology; \textbf{c,} Retina; \textbf{d,} Alzheimer MRI; \textbf{e,} Microscopy; \textbf{f,} Universe; \textbf{g,} Satellite; \textbf{h,} Describable Textures; \textbf{i,} Sketch Scene; \textbf{j,} Manga; \textbf{k,} Underwater; and \textbf{l,} Natural Scene. Detailed specifications for each dataset are provided in Supplementary Table S1.}

% ==================== Figure S4 ====================
\newpage
\begin{center}
    \includegraphics[width=0.9\textwidth]{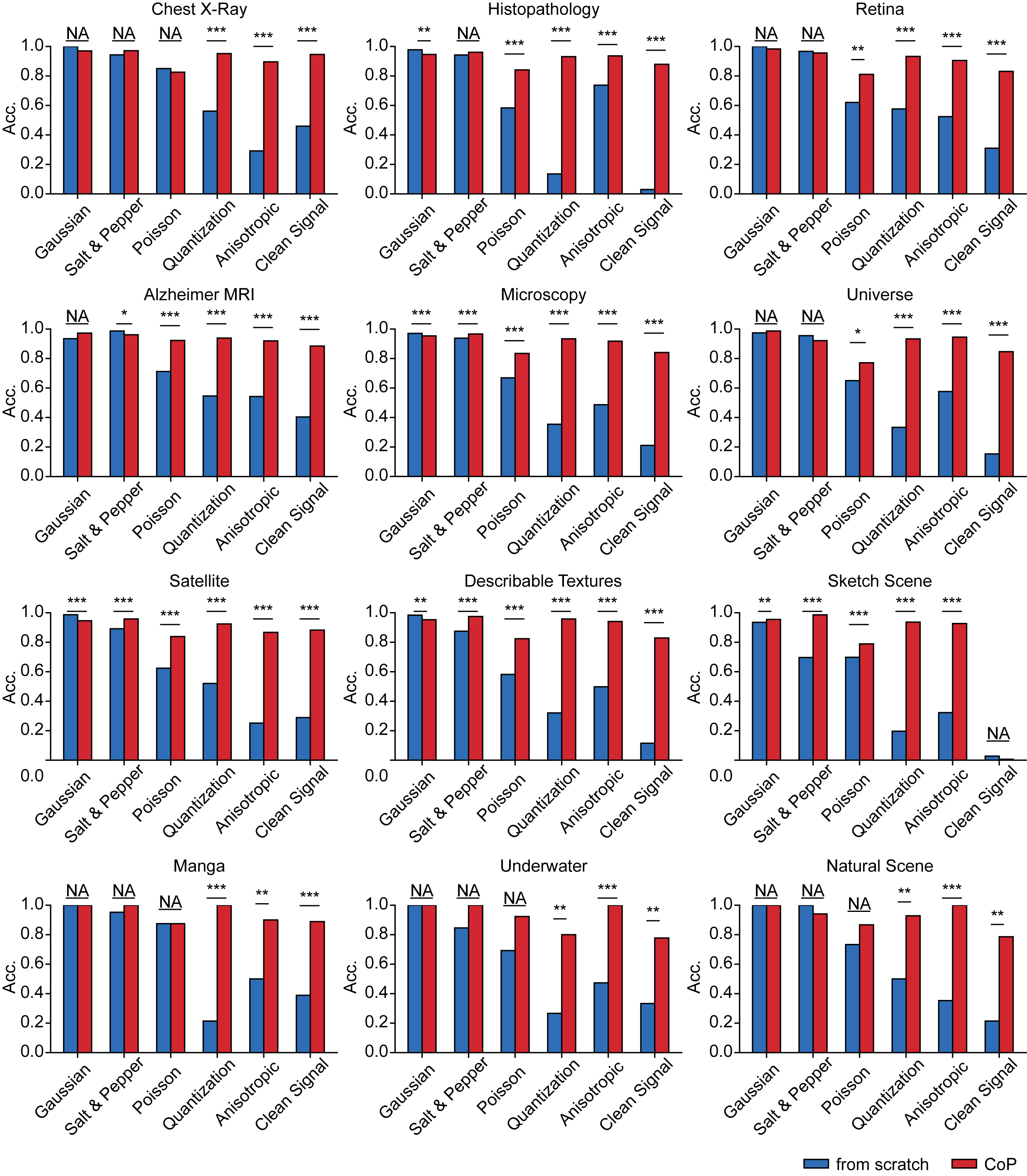}
\end{center}
\vspace{0.2em}
\captionof{figure}{\textbf{Comparisons of the accuracy of dominant noise category classification in cross modalities OOD data.}} 
\label{figS4} 
\vspace{0.2em} \noindent
{Performance comparison between the CoP foundation model (red bars) and the model trained from scratch (blue bars) on twelve distinct OOD datasets not seen during training. The bar charts quantify the classification accuracy (Acc.) for identifying the dominant noise type (Gaussian, Salt \& Pepper, Poisson, Quantization, Anisotropic) or detecting the Clean Signal. CoP demonstrates superior generalization capabilities, particularly for complex, non-stationary noise types such as Quantization and Anisotropic noise, as well as Clean Signal detection, where the baseline model frequently struggles. Statistical significance was determined using paired t-tests. Stars indicate significance levels: *$P < 0.05$, **$P < 0.01$, ***$P < 0.001$. NA indicates no statistically significant difference ($P > 0.05$), typically observed in simpler noise categories (e.g., Gaussian) where both models achieve near-perfect performance.}

% ==================== Figure S5 ====================
\newpage
\begin{center}
    \includegraphics[width=1\textwidth]{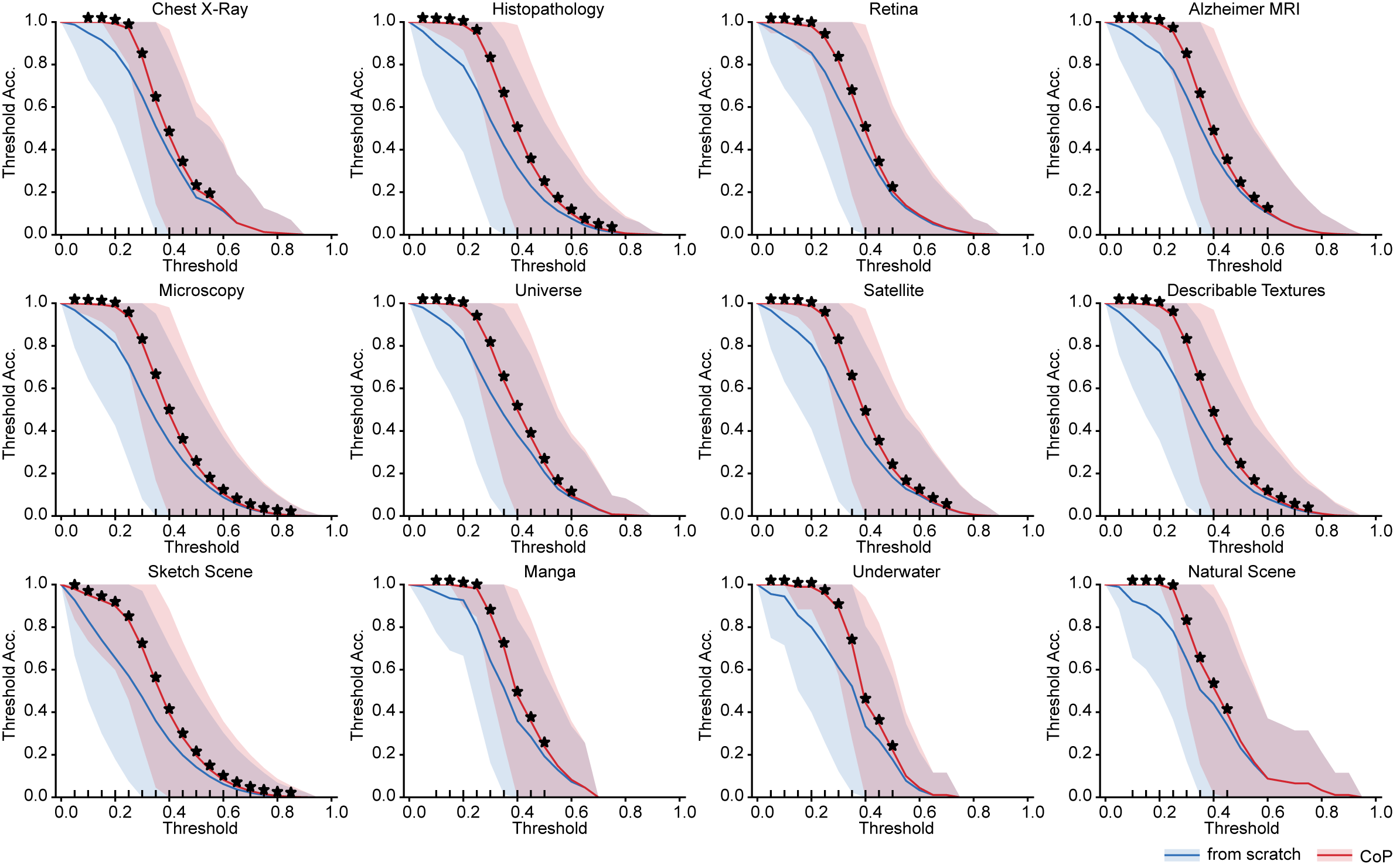}
\end{center}
\vspace{0.2em}
\captionof{figure}{\textbf{Comparisons of the threshold accuracy of multi-noise categories classification in cross modalities OOD data.}} 
\label{figS5} 
\vspace{0.2em} \noindent
{Quantitative evaluation of the robustness of noise classification predictions across varying decision thresholds. The curves plot the threshold accuracy ($y$-axis) against the decision threshold value ($x$-axis, range 0.0--1.0) for the CoP model (red line) versus the model trained from scratch (blue line) across the twelve OOD datasets. Shaded regions represent the standard deviation of the accuracy across test batches. The CoP model consistently maintains higher accuracy and slower degradation as the threshold increases compared to the model trained from scratch. This indicates that CoP not only predicts the correct noise labels but does so with higher confidence scores, exhibiting a more robust separation between signal and noise features in the latent space. Black stars ($\star$) denote thresholds where the performance difference between CoP and the baseline is statistically significant ($P < 0.001$, paired $t$-test).}

% ==================== Figure S6 ====================
\newpage
\begin{center}
    \includegraphics[width=0.75\textwidth]{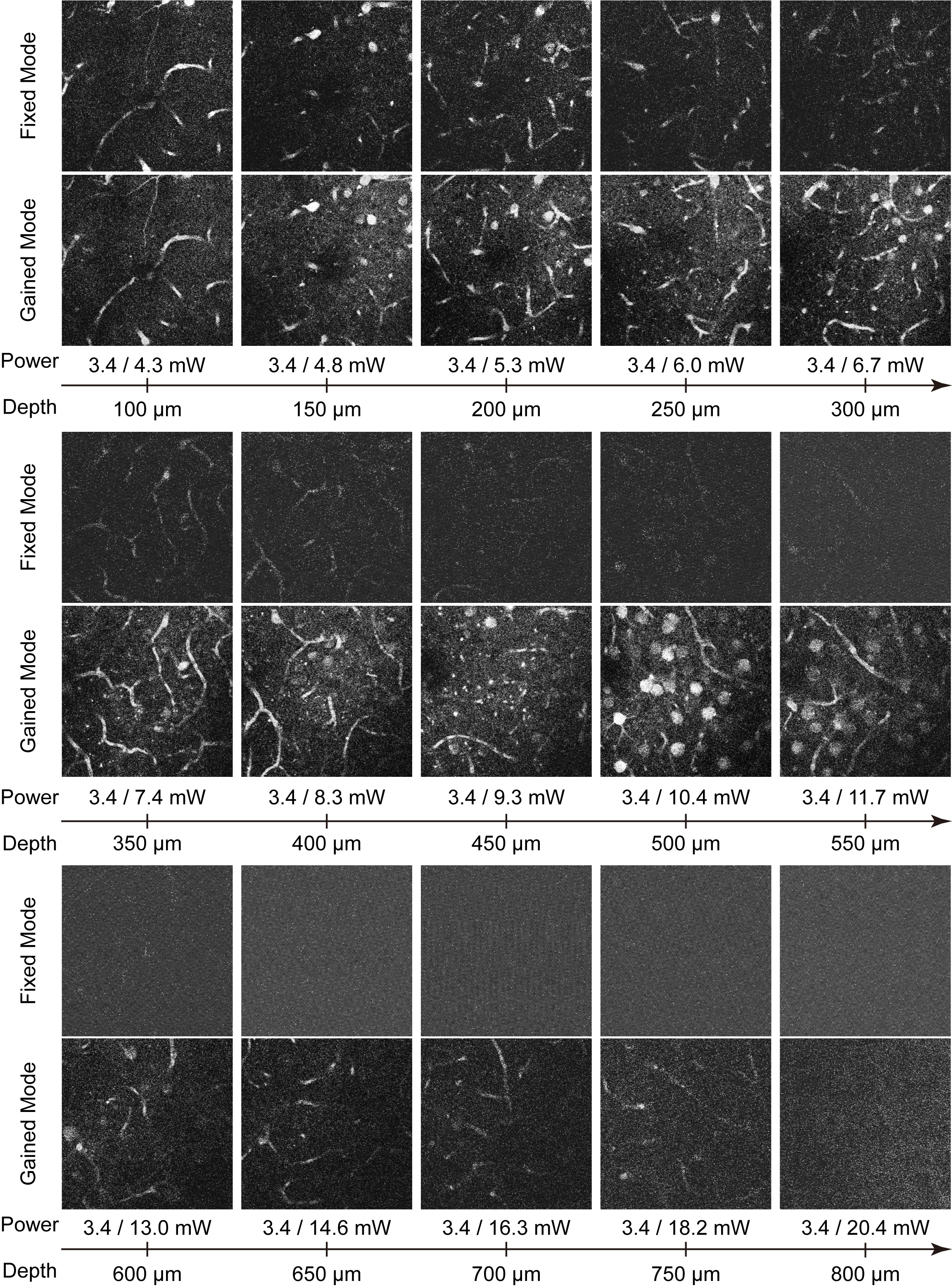}
\end{center}
\vspace{0.2em}
\captionof{figure}{\textbf{Visual comparison of imaging performance across depth under Fixed and Gained Excitation modes.}} 
\label{figS6} 
\vspace{0.2em} \noindent
{The panels display representative microscopy images captured from superficial ($100 \,\mu\mathrm{m}$) to deep ($800 \,\mu\mathrm{m}$) tissue layers. The fixed mode rows demonstrate progressive signal decay due to scattering, whereas the gained mode rows maintain visibility through adaptive power compensation. The corresponding excitation power levels are annotated below each column.}

% ==================== Table S1 ====================
\newpage
\begin{center}
    
    \renewcommand{\arraystretch}{1.3} 
    \setlength{\tabcolsep}{4pt} 
    
    \captionof{table}{\textbf{Details of the twelve Out-of-Distribution (OOD) datasets used for evaluation.}}
    \label{tabS1}
    \footnotesize 
    \begin{tabular}{|p{5cm}|l|r|p{8cm}|}
        \hline
        \rowcolor{gray!25} 
        \textbf{Dataset Name} & \textbf{Source Name} & \textbf{Image Count} & \textbf{Access URL} \\ 
        \hline
        
        Chest X-Ray~\citep{irvin2019chexpert} & CheXpert & 223,648 & \url{https://huggingface.co/datasets/danjacobellis/chexpert} \\
        \hline
        
        Histopathology~\citep{wei2021petri} & MHIST & 3,152 & \url{https://huggingface.co/datasets/mamunrobi35/mhist_binary} \\
        \hline
        
        Retina~\citep{yang2023medmnist} & RetinaMNIST & 1,600 & \url{https://medmnist.com/} \\
        \hline
        
        Alzheimer MRI~\citep{Falah2023Alzheimer} & Alzheimer MRI & 6,400 & \url{https://huggingface.co/datasets/Falah/Alzheimer_MRI} \\
        \hline
        
        Microscopy~\citep{kvriza2026Microscopy} & Microscopy Img. & 20,936 & \url{https://huggingface.co/datasets/kvriza8/microscopy_images} \\
        \hline
        
        Universe~\citep{liu2020faster} & FS Universe & 501 & \url{https://huggingface.co/datasets/huggan/few-shot-universe} \\
        \hline
        
        Satellite~\citep{helber2019eurosat} & EuroSAT & 43,200 & \url{https://huggingface.co/datasets/tanganke/eurosat} \\
        \hline
        
        Desc. Textures~\citep{cimpoi2014describing} & DTD & 5,640 & \url{https://www.robots.ox.ac.uk/~vgg/data/dtd/} \\
        \hline
        
        Sketch Scene~\citep{chowdhury2022fscoco} & Sketch-Scene & 9,999 & \url{https://huggingface.co/datasets/zoheb/sketch-scene} \\
        \hline
        
        Manga~\citep{matsui2017sketchbased} & Manga109 & 109 & \url{http://www.manga109.org/} \\
        \hline
        
        Underwater~\citep{li2020underwater} & UIEB & 890 & \url{https://li-chongyi.github.io/proj_benchmark.html} \\
        \hline
        
        Natural Scene~\citep{yang2010image} & T91 & 91 & \url{https://www.kaggle.com/datasets/ll01dm/t91-image-dataset} \\
        \hline
        
    \end{tabular}
\end{center}

% \clearpage
% \newpage
% \beginappendix
% \input{sec/appendix}

\end{document}